\documentclass[pmlr]{jmlr}

\usepackage{longtable}

\usepackage{booktabs}
\usepackage[load-configurations=version-1]{siunitx} 

\makeatletter
\def\set@curr@file#1{\def\@curr@file{#1}} 
\makeatother

\theorembodyfont{\upshape}
\theoremheaderfont{\scshape}
\theorempostheader{:}
\theoremsep{\newline}

\title[MedDialog: Two Large-scale Medical Dialogue Datasets]{MedDialog: Two Large-scale Medical Dialogue Datasets}

\author{\Name{Xuehai He}, \Name{Shu Chen}, \Name{Zeqian Ju}, \Name{Xiangyu Dong}, \Name{Hongchao Fang}, \Name{Sicheng Wang}, \Name{Yue Yang}, \Name{Jiaqi Zeng}, \Name{Ruisi Zhang}, \Name{Ruoyu Zhang}, \Name{Meng Zhou}, \Name{Penghui Zhu}, \Name{Pengtao Xie}\\
\addr UC San Diego\\
\Email{pengtaoxie2008@gmail.com}
       }

\begin{document}

\maketitle

\begin{abstract}
  Medical dialogue systems are promising in assisting in telemedicine to increase access to healthcare services, improve the quality of patient care, and reduce medical costs. To facilitate the research and development of medical dialogue systems, we build two  large-scale medical dialogue datasets: MedDialog-EN and MedDialog-CN. MedDialog-EN is an English dataset containing 0.3 million conversations between patients and doctors and 0.5 million utterances. MedDialog-CN is an Chinese dataset containing 1.1 million conversations and 4 million utterances.
  To our best knowledge, MedDialog-(EN,CN) are the largest medical dialogue datasets to date.  The dataset is available at \url{https://github.com/UCSD-AI4H/Medical-Dialogue-System}
\end{abstract}

\section{Introduction}
Telemedicine refers to the practice of delivering patient care remotely, where doctors provide medical consultations to patients using HIPAA compliant video-conferencing tools. As an important complement to traditional face-to-face medicine practiced physically in hospitals and clinics, telemedicine has a number of advantages. First, it increases access to care. For people living in medically under-served communities (e.g., rural areas) that are in shortage of clinicians, telemedicine enables them to receive faster and cheaper care compared with traveling over a long distance to visit a clinician. Second, it reduces healthcare cost. In a study\footnote{https://www.healthleadersmedia.com/clinical-care/cost-savings-telemedicine-estimated-19-120-patient-visit} by Jefferson Health, it is shown that diverting patients from emergency departments with telemedicine can save more than \$1,500 per visit. Third, telemedicine can improve quality of care. The study in \citep{pande2015leveraging} shows that telemedicine patients score lower for depression, anxiety, and stress, and have 38\% fewer hospital admissions. Other advantages include improving patient engagement and satisfaction, improving provider satisfaction, etc. Please refer to~\citep{wootton2017introduction} for a more comprehensive review.

While telemedicine is promising, it has several limitations. First, it puts additional burden to physicians. In additional to practicing face-to-face medicine which already makes physicians highly occupied, physicians need to provide remote consultations in telemedicine, which further increases the risk of physician burnout. Second, different from in-hospital patients, the progression of whose medical conditions can be easily tracked by clinicians, remote patients are difficult to track and monitor. To address such problems, there has been increasing research interests in developing artificial intelligence (AI) methods to assist in telemedicine. In particular, medical dialogue systems are being developed to server as ``virtual doctors". These ``virtual doctors" are aimed to interact with patients via natural dialogues, asking about the medical conditions and history of patients and providing clinical advice. They can also proactively reach out to patients to ask about the progression of patients' conditions and provide timely interventions accordingly.

\begin{figure}
\begin{center}
\includegraphics[width=\textwidth]{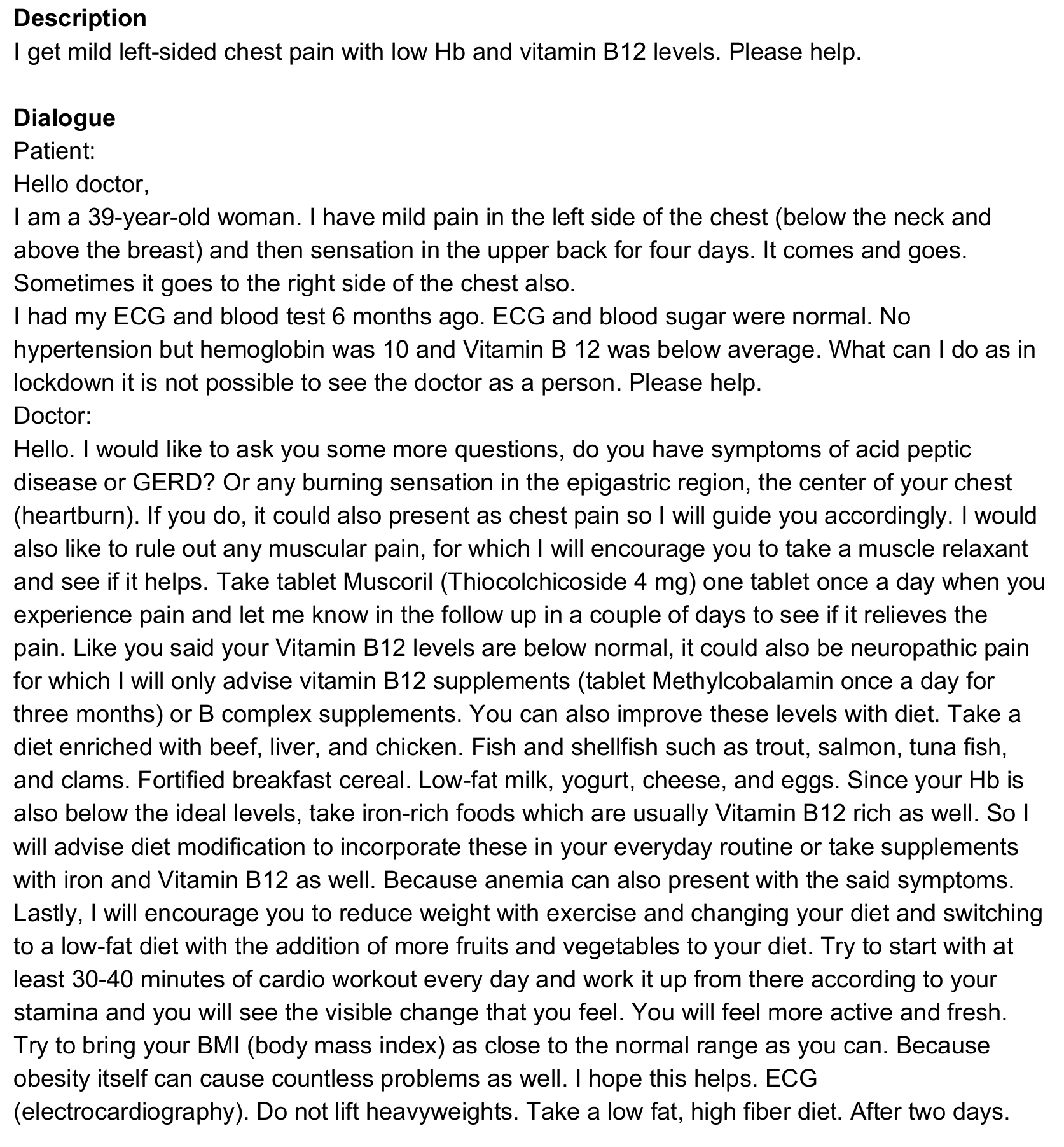}
\end{center}
\caption{An exemplar consultation, which includes (1) description of medical conditions of  the  patient, (2) dialogue between doctor and patient.}
\label{fig:eg_en}
\end{figure}

To build medical dialogue systems, a large collection of conversations between patients and doctors are needed as training data. Due to data privacy concerns, such data is very difficult to obtain. The existing medical dialogue datasets are limited in size or biased to certain diseases, which cannot adequately serve the purpose to train medical dialogue systems that can achieve doctor-level intelligence and cover all specialities in medicine. 

To address the limitations of existing datasets, we build two large-scale medical dialogue datasets: MedDialog-EN in English and MedDialog-CN in Chinese. MedDialog-EN contains 0.3 million patient-doctor consultations and 0.5 million utterances. MedDialog-CN contains 1.1 million consultations and 4 million utterances. Dialogs in these two datasets cover almost all specialities in medicine, ranging from internal medicine to family medicine and covers a wide spectrum of diseases, including cancer, pneumonia, etc. To our best knowledge, they are the largest medical dialogue datasets to date. The data is open to the public. 

\section{Datasets}

\subsection{MedDialog-EN}
The MedDialog-EN dataset contains 257,454 English consultations between patients and doctors. The total number of utterances is 514,908: 257,454 from doctors and 257,454 from patients. Each consultation consists of two parts: (1) description of patient's medical conditions; (2) conversation between patient and doctor. Figure~\ref{fig:eg_en} shows an exemplar consultation.  The data is crawled from iclinic.com\footnote{https://www.icliniq.com/} and  healthcaremagic.com\footnote{https://www.healthcaremagic.com/}, which are two online platforms of healthcare services, including 
symptom self-checker, video consultation, online chat with doctors, etc. 

The consultations cover 51 categories of communities including diabetes, elderly problems, pain management, etc. and 96 specialties including andrology, cardiology, nephrology, pharmacology, etc. The consultations are conducted from 2008 to 2020.

\begin{figure}[h]
\begin{center}
\includegraphics[width=\textwidth]{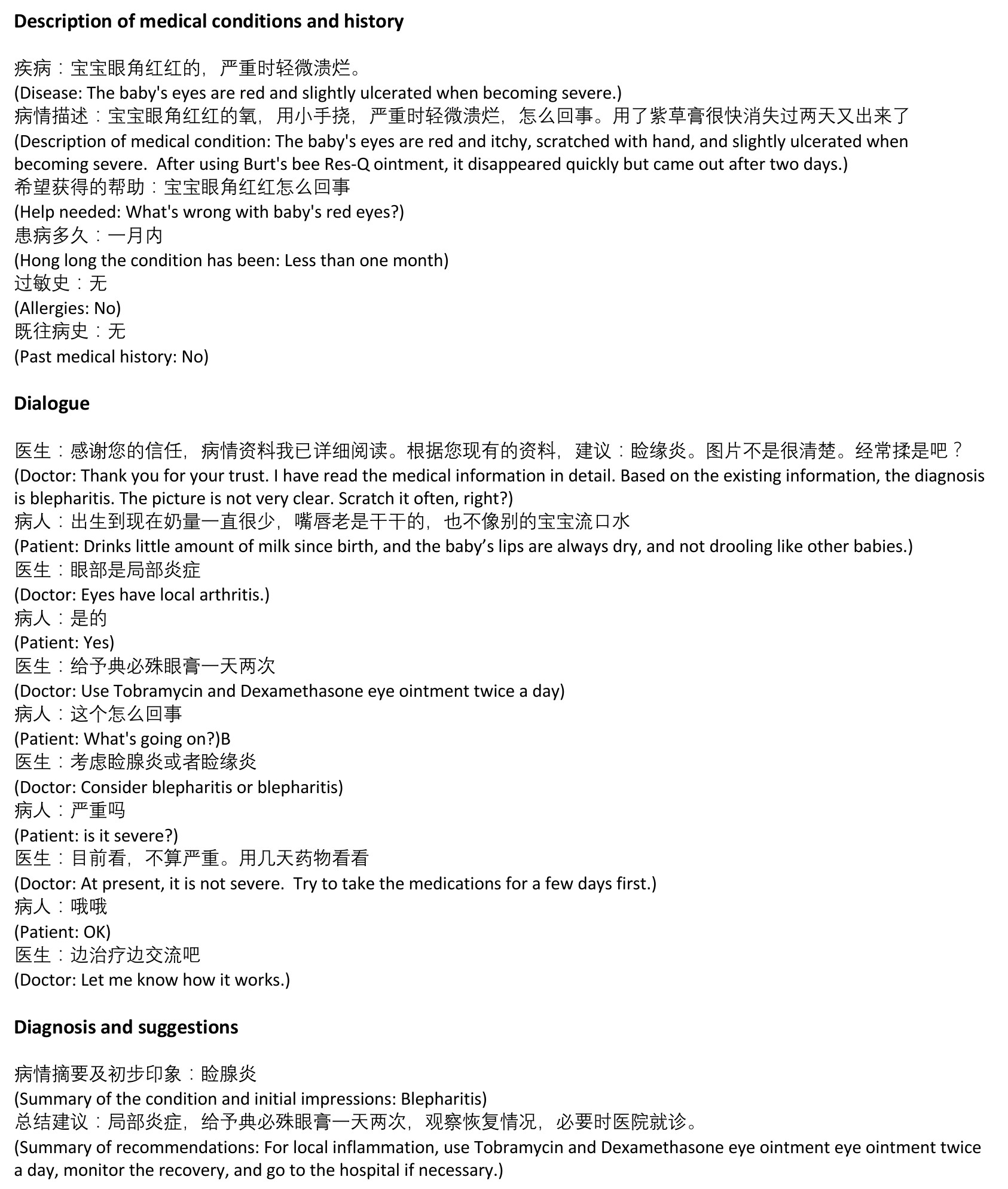}
\end{center}
\caption{An exemplar consultation, which includes (1) description of medical conditions and history of the patient, (2) dialogue between doctor and patient, and (3) diagnosis and treatment suggestions given by the doctor.}
\label{fig:eg}
\end{figure}

\subsection{MedDialog-CN}
The MedDialog-CN dataset contains 1,145,231 Chinese consultations between patients and doctors. The total number of utterances is 3,959,333: 2,179,008 from doctors and 1,780,325 from patients. Each consultation consists of three parts: (1) description of patient's medical condition and history; (2) conversation between patient and doctor; (3) (optional) diagnosis and treatment suggestions given by the doctor. In the description of patient's medical condition and history, the following fields are included: present disease, detailed description of present disease, what help is needed from the doctor, how long the disease has been, medications, allergies, and past disease. Figure~\ref{fig:eg} shows an exemplar consultation.   
In the conversation, there are cases that multiple consecutive utterances are from the same person (either doctor or patient) and these utterances were posted at different time points. If we combine consecutive utterances from the same person into a single one, there are 3,209,660 utterances: 1,981,844 from doctors and 1,227,816 from patients. The data is crawled from haodf.com\footnote{https://www.haodf.com/}, which is an online platform of healthcare services, including medical consultation, scheduling appointment with doctors, etc. 

The consultations cover 29 broad categories of specialties including internal medicine, pediatrics, dentistry, etc. and 172 fine-grained specialties including cardiology, neurology, gastroenterology, urology, etc. The consultations are conducted from 2010 to 2020.

\section{Advantages of our datasets}
To our best knowledge, MedDialog-EN and MedDialog-CN are the largest English and Chinese medical dialog dataset respectively. They have the following advantages. 

\begin{itemize}
    \item \textbf{Large number of conversations and utterances.} MedDialog-EN has about 0.3 million conversations and 0.5 million utterances. MedDialog-CN has about 1.1 million conversations and 4 million utterances.
    \item \textbf{Broad coverage of medical specialities.}  Consultations in MedDialog-EN are about 96 categories of specialties. Consultations in MedDialog-CN are about 29 broad categories of specialties and 172 fine-grained specialties. 
    \item \textbf{Diversity of the patients.} The patients in MedDialog-EN are from all over the world, with different nationalities, ethics, age, gender, occupation, education, income, etc.  The patients in MedDialog-CN are from 31 provincial-level administrative divisions in China, with different ethics, age, gender, occupation, education, income, etc. Such diversity greatly minimizes population biases in these two datasets. 
\end{itemize}

\section{Related Works}
Table~\ref{tb:cmp} shows a comparison of our dataset with several other medical dialogue datasets. The number of dialogs and diseases in our dataset are both much larger than those in other datasets. \begin{table}[htbp]
  \centering 
  \small
 \begin{tabular}{l|c|c}
 \hline
Dataset & \#dialogs &  \#diseases  \\
\hline
Muzhi~\citep{wei2018task} & 710 & 4 \\
Dxy~\citep{xu2019end} & 527 & 5\\
COVID-EN~\citep{yang2020generation}    &  603  &   1\\
COVID-CN~\citep{yang2020generation}     & 1,088   &   1\\
\hline
\hline
MedDialog-EN & 257,454&  96\\
MedDialog-CN & 3,407,494 & 172 \\
\hline
  \end{tabular}
  \caption{Comparison with other datasets.}
  \label{tb:cmp} 
  \vspace{-0.3cm}
\end{table}

\section{Conclusions}
To facilitate the research and development of medical dialogue systems that can potentially assist in telemedicine, we build two large-scale medical dialogue datasets. MedDialog-EN contains 0.3 million conversations between patients and doctors and 0.5 million utterances. MedDialog-CN contains 1.1 million conversations and 4 million utterances. The datasets are publicly available and are continuously growing.

\bibliographystyle{unsrt}
\bibliography{release}

\end{document}